\documentclass{article}
\usepackage{amsmath}
\usepackage{float}
\usepackage{hyperref} 
\usepackage[left=3cm,right=2cm,top=2cm,bottom=2cm,bindingoffset=0cm]{geometry}
\usepackage{graphicx}

\graphicspath{ {./images/} }

\newcommand{\norm}[1]{\lVert#1\rVert}
\newcommand{\argmax}{\mathop{\mathrm{argmax}}\limits} 

\begin{document}

\title{Sets of autoencoders with shared latent spaces}
\author{Vasily Morzhakov}
\maketitle


\begin{abstract}
Autoencoders receive latent models of input data. It was shown in recent works that they also estimate probability density functions of the input. This fact makes using the Bayesian decision theory possible. If we obtain latent models of input data for each class or for some points in the space of parameters in a parameter estimation task, we are able to estimate likelihood functions for those classes or points in parameter space. We show how the set of autoencoders solves the recognition problem. Each autoencoder describes its own model or context, a latent vector that presents input data in the latent space may be called “treatment” in its context. Sharing latent spaces of autoencoders gives a very important property that is the ability to separate treatment and context where the input information is treated through the set of autoencoders. There are two remarkable and most valuable results of this work: a mechanism that shows a possible way of forming abstract concepts and a way of reducing dataset’s size during training. These results are confirmed by tests presented in the article.
\end{abstract}


\section{Introduction}

	Tasks of classification and parameter estimation have strict probabilistic background that is Bayesian Decision Theory and a Bayesian estimator. Originally, Artificial Neural Networks (ANN) were based on probabilistic models, but nowadays the best applied solutions usually don’t estimate distributions of data, that behavior brings with it  some fundamental problems that can’t be eliminated without changing the approach:
\begin{enumerate} 
	\item Adversarial patches [6] change the output of Convolutional Neural Networks. Uncontrollable back- propagation optimization of millions of parameters doesn’t allow networks to be stable even in close proximity of points presented in training sets.
	\item It is extremely difficult to locate the source of a problem in trained Neural Networks (NN) and eliminate it by partial retraining. It works as a ”black box”.
	\item Multiple interpretations are not allowed, also the statistical nature of input data is ignored.
\end{enumerate} 

\begin{figure}[h]
\center{\includegraphics[height=4cm]{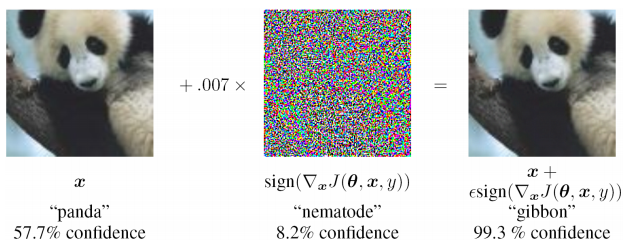}}
\caption{An example of the adversarial attack from the article [6]}
\label{fig:image}
\end{figure}

\begin{figure}[h]
\center{\includegraphics[height=6cm]{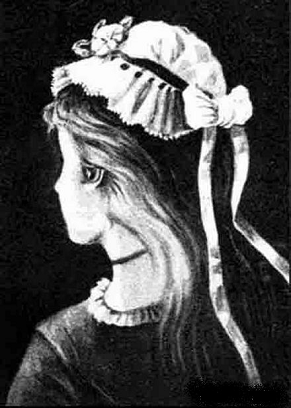}}
\caption{An example of multiple interpretations in vision: "old or young lady"}
\label{fig:image}
\end{figure}

\section{Autoencoders and a probability density function}

The probability density function (PDF) plays a very important role in the Bayesian decision theory. It’s required to know a PDF of input data for calculation a posteriori risk in choosing the best decision in classification tasks.

It turned out that autoencoders are very suitable for estimating of the PDF. It could be explained in the following way: the training data set defines the density of input data, meaning the more the training samples are placed around a point in the input space, the better the autoencoder will reconstruct the input there. Besides, there is a latent vector in the bottleneck of the autoencoder, if an input vector is projected to the area in the latent space that was not involved previously in training, then this input vector is unlikely.

There are a few works where the connection between autoencoders and PDFs is proved theoretically: Alain, G. and Bengio, Y. What regularized autoencoders learn from the data generating distribution. 2013. [1] ,Kamyshanska, H. 2013. On autoencoder scoring [2],Daniel Jiwoong Im, Mohamed Ishmael Belghazi, Roland Memisevic. 2016. Conservativeness of Untied Auto-Encoders [3].

In this article another approach to estimation of a PDF will be shown that gives clues to construction of sets of autoencoders with the shared latent space.

\section{Single autoencoder}

We are considering an arbitrary autoencoder $x^*=f(g(x))$, where $g(x)$ - encoder, $f(z)$ - decoder. The encoder projects the input space to the corresponding latent space.\\

A probability density function for $x \in X$ and $z \in Z$ is equal to: 

\begin{equation}
p(x) = \int_{z}  p (x|z)p(z)dz
\end{equation}

Our goal is to obtain a relationship between $p(x)$ and $p(z)$, because p(z) will be an issue related to this research in the following.\\

For simplicity, let's assume that the input vector $x$ is presented as $x=f(g(x))+n$  after autoencoder training, where $n$ is gaussian noise, there is a ”latent model” that we were able to receive, for example, by training a multiple-layer neural network.
Then the noise distribution $n=f(z)-x$ with the deviation $\sigma$ is:\\
\begin{equation}
p(n)=const\times exp( -\frac{(x-f(z))^T (x-f(z))}{2\sigma^2} )=p(x|z)
\end{equation}

$(x-f(z))^T (x-f(z))$ is the distance between $x$ and its reprojection through the latent space back to $X$.This distance reaches its minimum value at some point $z^*$.
Partial derivatives of the exponent’s argument in  the equation (2) will be zero in the direction $z_i$, where $z_i$ are axes in $Z$. \\
\begin{equation*}
0 = \frac{\partial f(z^*)}{\partial z_i}^T(x-f(z^*))+(x-f(z^*))^T\frac{\partial f(z^*)}{\partial z_i}
\end{equation*}
$\frac{\partial f(z^*)}{\partial z_i}^T(x-f(z^*))$ is a scalar, then:\\
\begin{equation}
0 = \frac{\partial f(z^*)}{\partial z_i}^T(x-f(z^*))
\end{equation}\\

Choosing the point where the distance $\norm{x-f(z)}$ has its minimum value is founded on the weights’ optimization of the autoencoder. While training, least square error (L2) loss between input and output is minimized for all train samples: 
\begin{equation*}
\min\limits_{\theta, \forall x \in X train}\norm{x-f_\theta(g_\theta(x))}
\end{equation*}
$\theta$ - autoencoder's weights.\\

After success autoencoder training, varying weights brings $g(x)$ to the $z^*$, and we can consider it as estimation.\\

We also can present $f(z)$ through the first Taylor member around $z^*$  in (2) 
\begin{equation*}
f(z)=f(z^*)+\nabla f(z^*)(z-z^*)+o((z-z^*))
\end{equation*}

so the equation (2) is now
\begin{equation*}
p(x|z) \approx const\times exp( -\frac{((x-f(z^*))-\nabla f(z^*)(z-z^*))^T ((x-f(z^*))-\nabla f(z^*)(z-z^*))}{2\sigma^2} )=
\end{equation*}
\begin{equation*}
= const\times exp(-\frac{(x-f(z^*))^T(x-f(z^*))}{2\sigma^2})exp(-\frac{(\nabla f(z^*)(z-z^*))^T(\nabla f(z^*)(z-z^*))}{2\sigma^2}) \times
\end{equation*}
\begin{equation*}
\times exp(-\frac{( \nabla f(z^*)^T(x-f(z^*))+(x-f(z^*))^T\nabla f(z^*))(z-z^*)}{2\sigma^2})
\end{equation*}
Note that the last multiplier is equal to 1 according to the equation (3). The first multiplier doesn't depend on $z$ and it can be brought outside the integral sign. 
Another assumption we will make is that $p(z)$ is a smooth function and it can be replaced by $p(z^*)$ around $z^*$. \\

After making all assumptions, the integral (1) can be estimated:
\begin{equation*}
p(x) = const\times p(z^*)exp(- \frac{(x-f(z^*))^T(x-f(z^*))}{2\sigma^2}) \int_{z}exp(-(z-z^*)^TW(x)^TW(x)(z-z^*)) dz,   z^*=g(x)
\end{equation*}
where $W(x)=\frac{\nabla f(z^*)}{\sigma}, z^* = g(x)$.\\

The last intergal is the n-dimensional Euler-Poisson integral:
\begin{equation*}
 \int_{z}exp(-\frac{(z-z^*)^TW(x)^TW(x)(z-z^*)}{2}) dz=\sqrt{\frac{1}{det(W(x)^TW(x)/2\pi)}}
\end{equation*}

Finally, the distribution $p(x)$ has the following approximation:
\begin{equation}
p(x) = const\times exp(- \frac{(x-f(z^*))^T(x-f(z^*))}{2\sigma^2})p(z^*)\sqrt{\frac{1}{det(W(x)^TW(x)/2\pi)}},   z^*=g(x)
\end{equation}

We have shown that the input data distribution $p(x)$ can be estimated by multiplication of three factors:
\begin{enumerate} 
	\item The distance between the input vector and its reconstruction
	\item The distribution $p(z)$ at the projected point $z^*=g(x)$
	\item The integral value, that is calculated directly from autoencoder's weights
\end{enumerate} 
Despite all assumption we have made the result still makes sense and has the right to exist. 

\section{Set of autoencoders}

The idea of using a set of autoencoders for choosing the most likely solution in classification is not a new one [2]. As we showed above, the distribution $p_{class}(x)$ can be estimated by a trained autoencoder for each class. $p_{class}(x)$ corresponds to the likelihood function. Then, according to the Bayes' decision theory, we simply choose
$\argmax_{class} p_{class}(x)$. \\

 Thanks to (4) we estimate $\log{p_{class}(x)}$ for each class:\\
\begin{equation}
\log{p(x)} = const1+ const2\times(- (x-f(z^*))^T(x-f(z^*)))+\log{( p(z^*)\sqrt{\frac{1}{det(W(x)^TW(x)/2\pi)}})},   z^*=g(x)
\end{equation}

It is possible to form vectors from all terms in the equation (5) for each autoencoder and tune $const1,const2$ by the back propogation, as it was proposed in "On autoencoders scoring" [2].

Tests on the MNIST dataset were run. It was important to compare a set of autoencoders with the classic classification fully-connected network. Only the first term in (5) was used, because otherwise the estimation of p(z) would be required.

The table (1) presents architectures of both nets:

\begin{table}[H]
\begin{center}
    \begin{tabular}{ | l | p{5cm} | p{5cm} |}
    \hline
    \textbf{Layer} & \textbf{Autoencoder} & \textbf{Classification NN} \\ \hline
    1 & 2048 fully-connected neurons, relu activation & 2048 fully-connected neurons, relu activation \\ \hline
    2 & 16 fully-connected neurons, no activation (latent code) & 10 fully-connected neurons, softmax activation  \\    \hline
    3 & 2048 fully-connected neurons, relu activation, weights are tied to the layer 2 weights &\\    \hline
    5 & 784 fully-connected neurons, sigmoid activation, weights are tied to the layer 1 weights &\\    \hline
    \end{tabular}
\end{center}
\caption{Architectures of autoencoders and a classification ANN}
\end{table}

The Artificial Neural Network (ANN) was trained with the crossentropy loss.\\

Autoencoders' accuracy: 98.6\%
Classification NN: 98.4\%.\\

Of course, Convolutional NNs gain higher accuracy, but the task was just to compare two approaches: classification NNs with a hidden layer and estimating PDFs with autoencoders.

One more very important consequence of using autoencoders is that it's possible to detect anomalies: if reconstruction of input data was wrong or projection to the latent space had not appeared previously then the input was an anomaly.

\section{Sharing the latent space}

The most novel part of this research is sharing the latent space between autoencoders. Training autoencoder we are trying to receive "treatment" of the input data, that describes it sufficiently for the reconstruction. In the case of using a set of autoencoders it may be that this treatment can be the same for different autoencoders. For example, in a computer vision task we can define each autoencoder for a corresponding orientation. Then a point in the latent space of an autoencoder will define the properties of an object in the field of view. Those properties are actually the same for all orientations, then its presentation in latent spaces of different autoencoders should be the same. There are no obvious consequences of this proposal, but it will help to improve the estimation of p(z) in (4) and receive more abstract concepts from the input data. The distribution  p(z) becomes shared for all autoencoders and all samples we have in all contexts are projected into the same space Z. Also this will provide transfer of samples into different contests that allows one-shot learning. 

\section{Substitutability of classification tasks  and parameter estimation tasks}
We still hasn't discussed how we choose the number of autoencoders of the number of contexts. It seems there is no strict rule. We can train an autoencoder for a class or for a value in some parameter space. To clarify it an example is needed. Let's consider the task of face recognition:

Input images are human faces, then two different approaches are allowed:
1) context is face orientation. In this case reconstruction of an input image requires "treatment" that is a face-identification code. During training we will need to show the same face from different directions, "freezing" its latent code.
2) context is face's identity. In this case reconstruction of an input image requires face's orientation. During training we will show different faces from the same direction. 

An Optimal Bayesian decision will be chosen in regard to face's orientation in the first case, and in regard to face's identity in the second. 

\section{Cross-training procedure in training of the set of autoencoders}

Besides regular autoencoders' training we need to make the latent shared between them. In order to receive tied latent spaces the same essense should be demonstrated in different contexts. For example, speaking of MNIST images, cross-training could be presented schematically in figure 3.

\begin{figure}[h]
\center{\includegraphics[height=6cm]{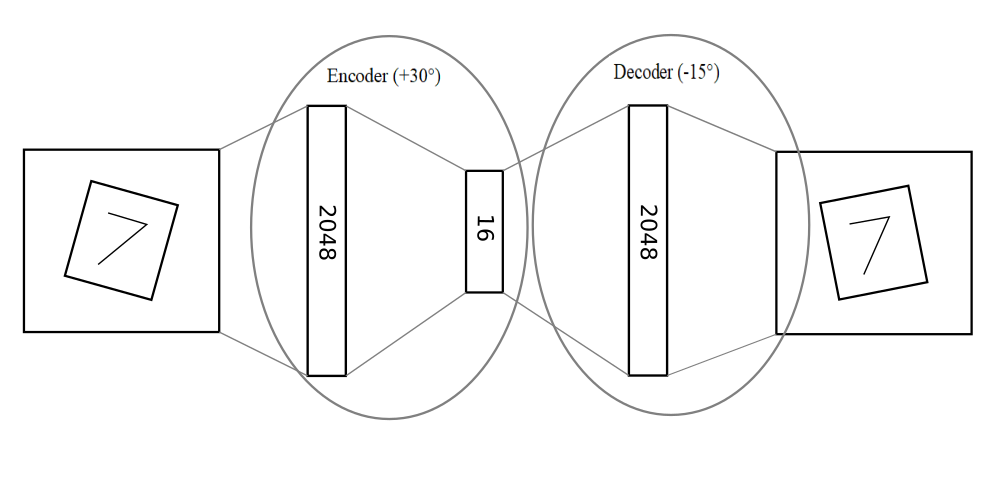}}
\caption{Scheme of cross-training}
\label{fig:image}
\end{figure}

Firstly, the encoder of an autoencoder encode an image, then the received latent code is decoded by the decoder of another autoencoder.This pipeline translates an image from one context to another.

Regular self-training steps for each autoencoder alternate with cross-training steps. At the end all autoencoders will receive shared latent space or "treatment" will be tied in different contexts.

\section{Training sample}

Let's consider a sample that is based on the MNIST dataset. It would demonstrate principles of training sets of autoencoders with the shared latent space. Finally, in this sample two effects will be shown: forming an absract concept of "cube" and one-shot learning.\\

Images of digits are sprayed on cube edges.\\
\begin{figure}[h]
\center{\includegraphics[height=6cm]{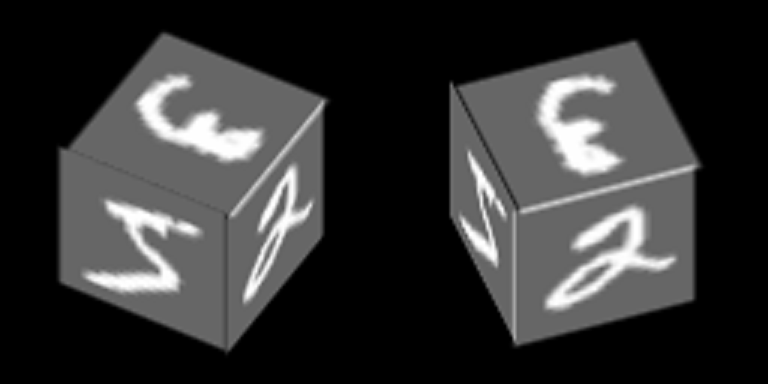}}
\caption{Cube samples}
\label{fig:image}
\end{figure}

Autoencoders will reconstruct edges in this sample, contexts are orientations  of those edges.\\

Figure 5 shows a "zero" digit in 100 orientational contexts, the first 34 of them correspond to different orientations of a side edge and the other 76 to orientations of the top edge.\\

\begin{figure}[H]
\centering
\includegraphics[height=11cm]{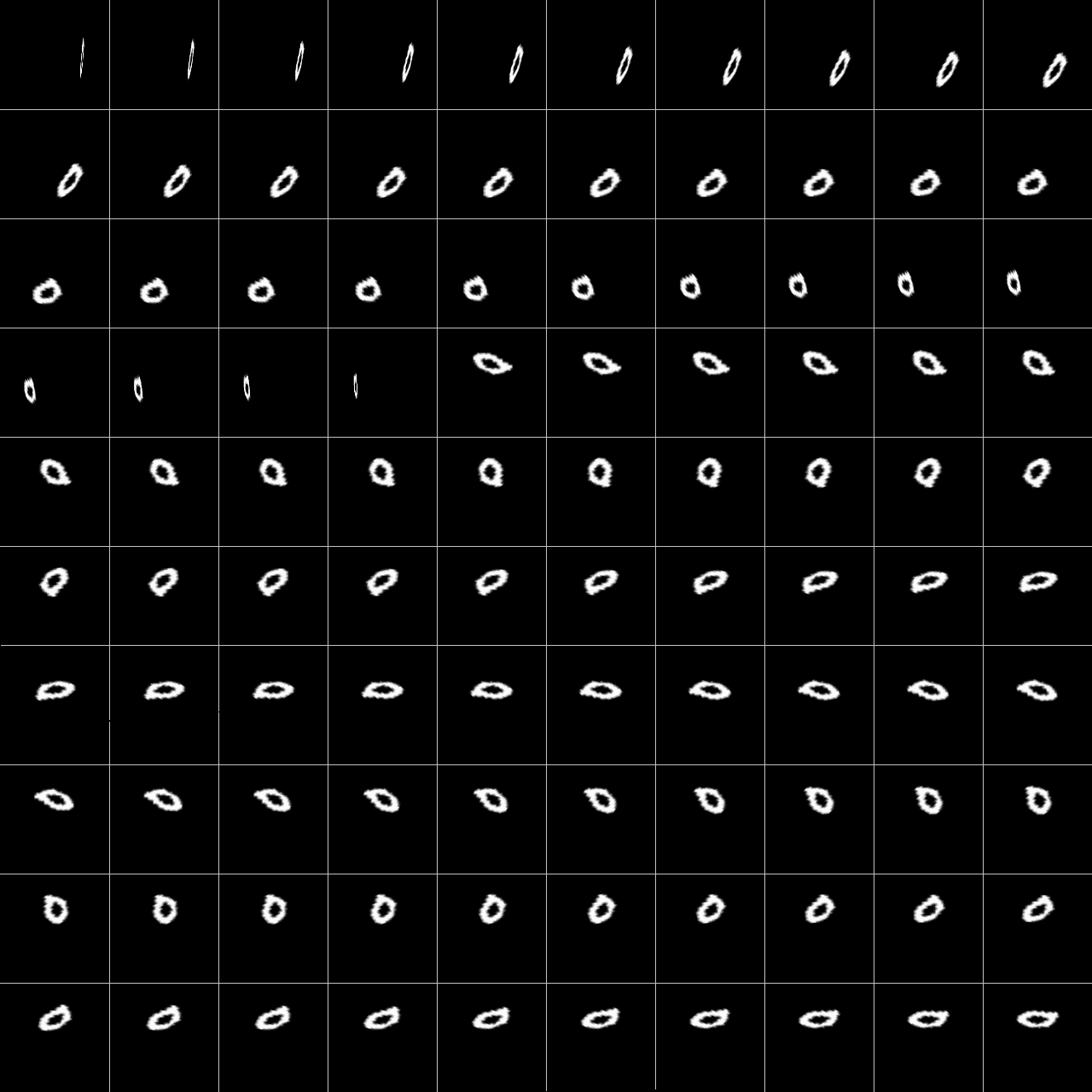}
\caption{The same digit is in different contexts}
\label{fig:image}
\end{figure}

It is proposed that each of these 100 images have the same "treatment" or the same value of their latent vector. Arbitrary pairs of them are used for cross-training. \\
Following the proposed cross-training approach autoencoders were successfully trained. A latent code of one autoencoders may be decoded by another and sensible translation will be reconstructed. 

Figure 6 shows the result of context transfering. An input image is encoded by the encoder 10th and decoded by others autoencoders:

\begin{figure}[H]
\centering
\includegraphics[width=16cm]{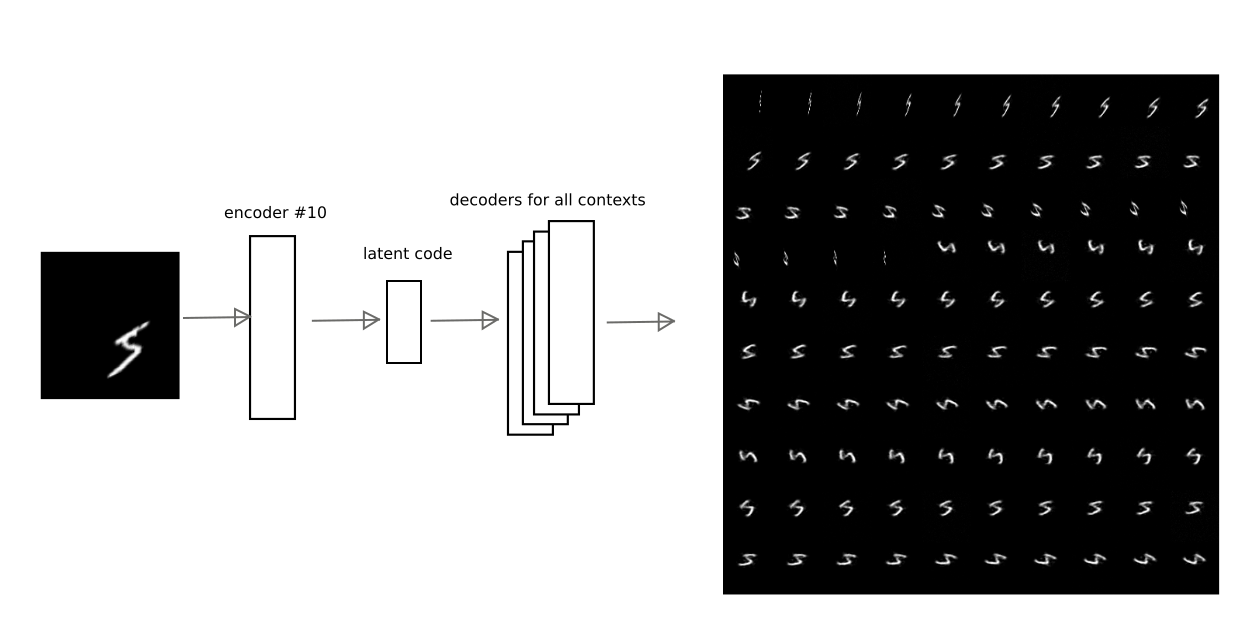}
\caption{Context transfering}
\label{fig:image}
\end{figure}

Thus, a latent representation (treatment) of an input image may be reconstructed in any of trained contexts.

\section{Masking in reconstruction}

In the equation (4) deviation $\sigma$ appears and it was chosen as a constant for all components of an input vector. However, if some components have no connection with latent models, then this deviation will be significantly higher for those components. $\sigma$ is in the denominator, it means the higher deviation in residual is the less contribution to probability estimation it has. It can be considered as partly masking of inputs.\\

In the shown example of cube edges masks are obvious.\\
The approach was simplified a bit: residual between an input vector and its reconstruction is multiplied by the masks. In general, more precise deviation estimation is required.

\begin{figure}[H]
\centering
\includegraphics[height=11cm]{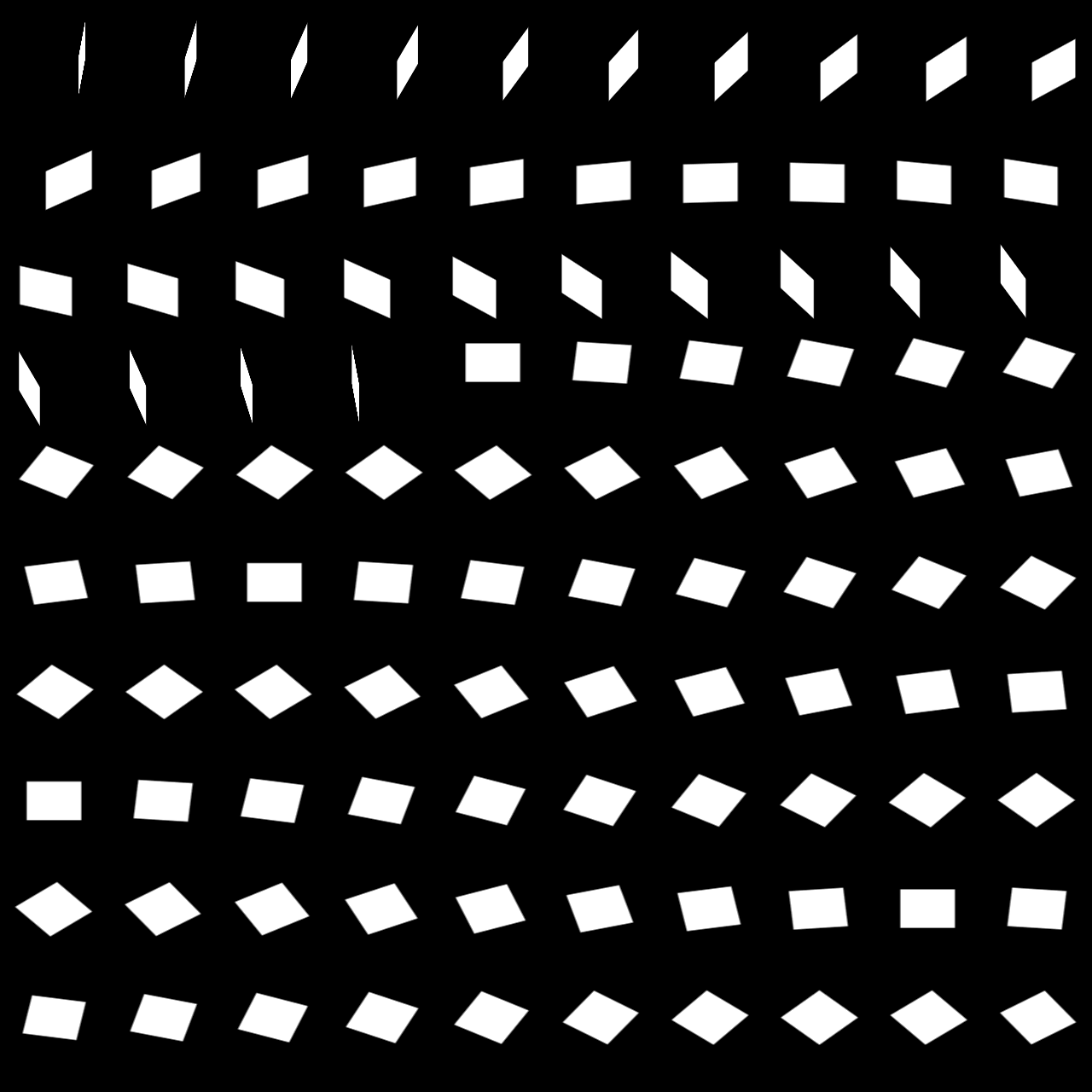}
\caption{Masks for cube edges}
\label{fig:image}
\end{figure}

\section{Idea of on-shot learning}

In some cases it's reasonable to operate with latent codes ignoring the context. This allows us to recognize a pattern in different contexts if it was demonstrated only in one of them. For one-shot learning it's important to show a new pattern for recognition. For the purpose of testing let's try to distinguish two different patterns that were not presented in the MNIST dataset:

\begin{figure}[H]
\centering
\includegraphics[height=6cm]{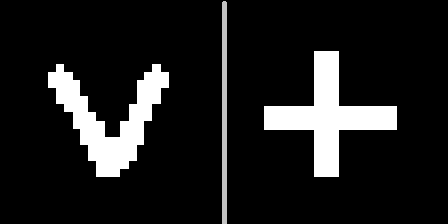}
\caption{Two new patterns}
\label{fig:image}
\end{figure}

These signs will be shown in one orientation, after that a hyperplane between two points that correspond to treatments of two signs should separate these signs in other orientations. \\
It's important to notice that autoencoders will not be trained on new signs. Thanks to variety of digits in MNIST it is possible to project new signs in the latent space and project it back to the image space in different orientations. For example, the sign V will be decoded in the following way:

\begin{figure}[H]
\centering
\includegraphics[height=7cm]{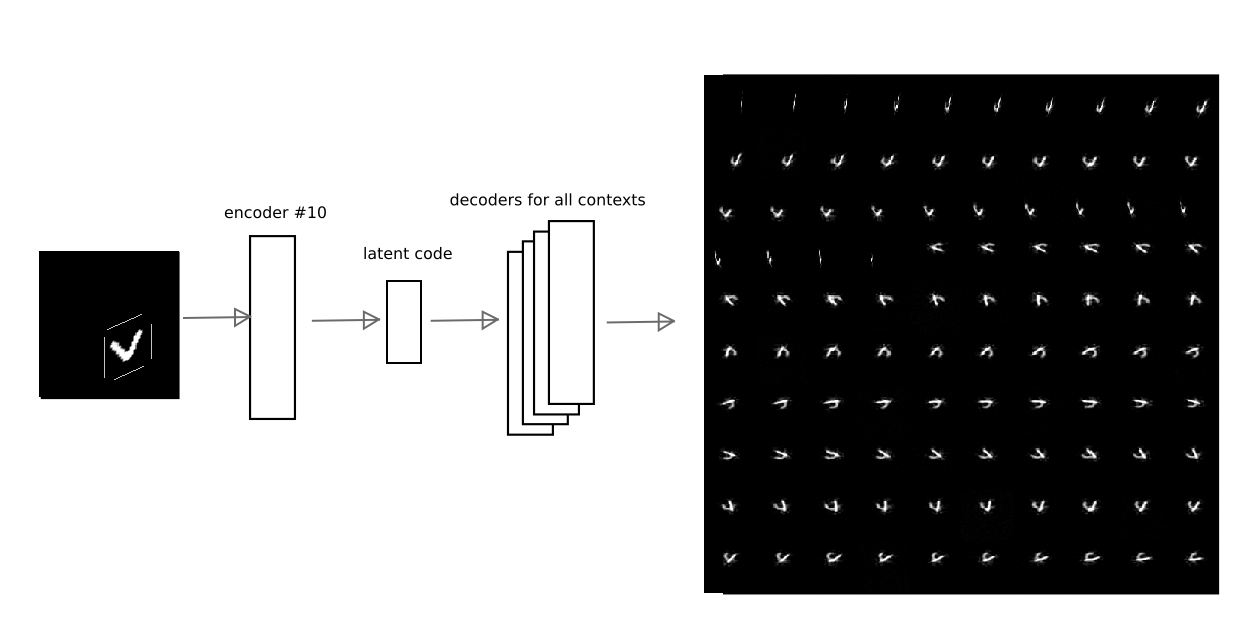}
\caption{V sign reconstructions in different contexts} 
\label{fig:image}
\end{figure}
 
Finnaly, successfull separation was obtained. Showing just one sample per class allowed us to recognize signes in all 100 contexts.

\begin{figure}[H]
\centering
\includegraphics[height=10cm]{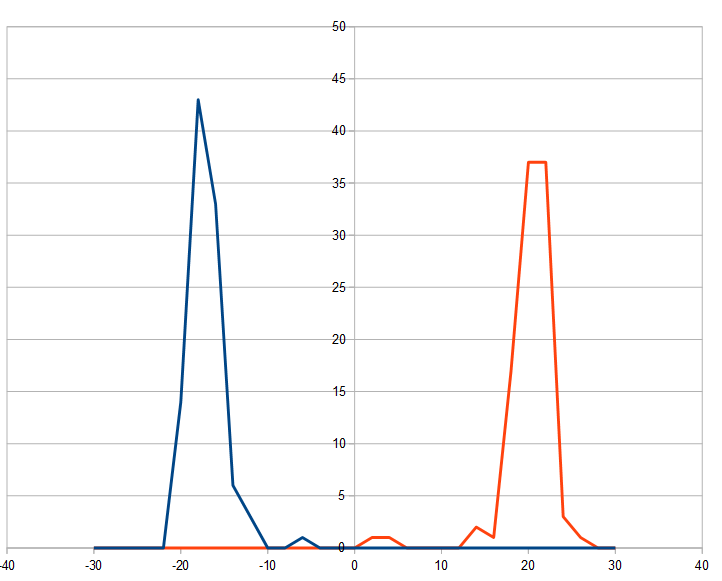}
\caption{Distance distribution}
\label{fig:image}
\end{figure}

It's also possible to visualize the result on separate images:

\begin{figure}[H]
\centering
\includegraphics[height=6cm]{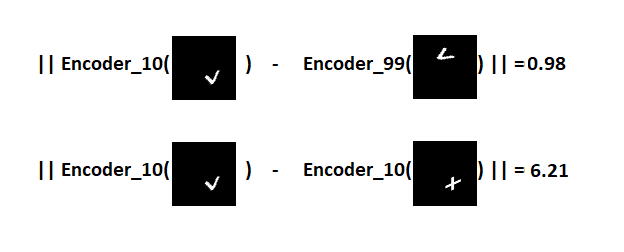}
\caption{Distance distribution}
\label{fig:image}
\end{figure}

This approach is similar to the ideas of transfer-learning for one-shot learning proposed in [7].

\section{Idea of abstract concepts}

The latent representation (treatment) can be ignored in some cases, so just contexts’ likelihood may be passed to higher processing level.
Let’s consider samples of contexts’ likelihood vectors for cubes in the same orientation but with different signs sprayed on their edges (digits 5 and 9). These vectors are shown as 10x10 maps:\\

\begin{figure}[H]
\center{\includegraphics[width=16cm]{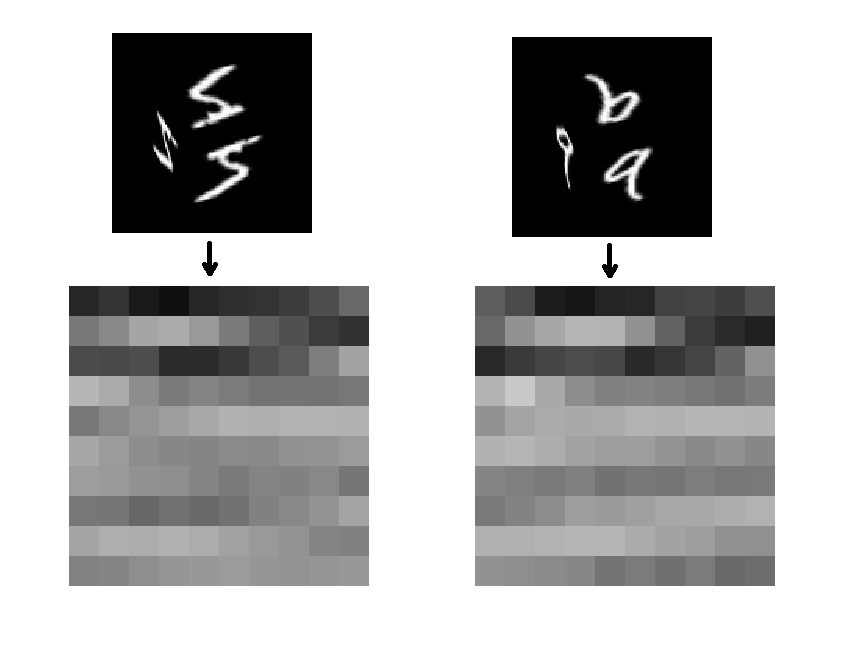}}
\caption{Likelihood context maps}
\label{fig:image}
\end{figure}

Maps are almost identical despite the different textures on cube edges. It means that likelihood vectors allows us to formulate a new concept that is 3D cube. So we need to train one more level with just one autoencoder that will be responsible for reconstructing different orientations of 3D cubes. Put simply, the next level will force the 3D cube to rotate.\\
Thanks to ignoring the latent code it's possible to train a new concept showing just one digit on cube edges. So, the experiment will be constructed in the following way:
\begin{enumerate} 
\item The training dataset contains images of cubes with rotation angle from 0 to 90 degrees. On all edges the same digit 5 is sprayed. 
\item Likelihood contexts’ vector is passed to the next level where only one autoencoder responsible for the cube model is placed.
\item Then back projection is being implemented. For points in the latent space in the second level it is possible to decode reconstructed likelihood contexts’ vector, add latent codes in the first level and reconstruct an image with a rotated cube on it.

\end{enumerate}

The training dataset consisted of 54321 images, here are some examples:\\
\begin{figure}[h]
\center{\includegraphics[height=1.4cm]{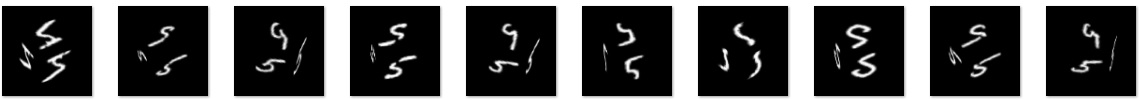}}
\caption{Cubes that were rotated from 0 to 90 degrees}
\label{fig:image}
\end{figure}\\
Cubes rotated from 0 to 90 degrees.
The Autoencoder on the second level had only one component in the latent vector, because it was known that we were dealing with only one degree of freedom in rotation. After training this component could be varied from 0 to 1 and decode the likelihood contexts’ vector:\\
\begin{figure}[h]
\center{\includegraphics[height=1.4cm]{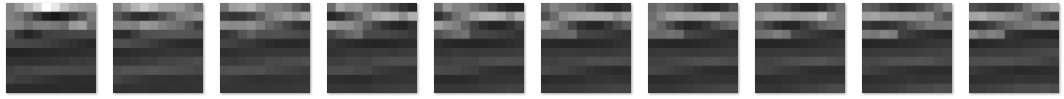}}
\caption{Likelihood context map's varity}
\label{fig:image}
\end{figure}\\

Then, this vector is translated back to the first level, local maximums are chosen and an arbitrary latent code of edge’s texture is taken. Let it be a code of a “3” digit. Changing the component of latent code on the second level we receive the following reconstructed images:\\
\begin{figure}[h]
\center{\includegraphics[height=1.4cm]{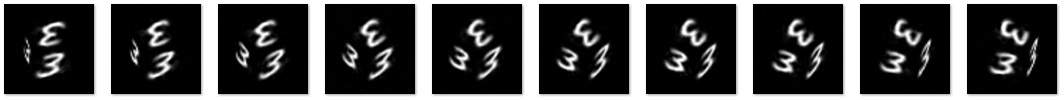}}
\caption{Back projection with the changed latent code}
\label{fig:image}
\end{figure}\\
Also, we can generate (imagine) a new cube that was not presented in training data with a “V” sign on edges:\\
\begin{figure}[h]
\center{\includegraphics[height=1.4cm]{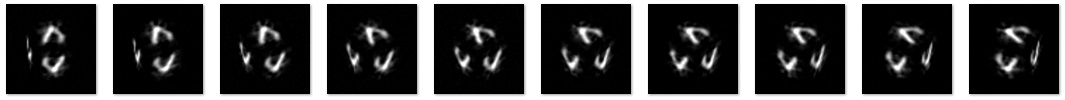}}
\caption{Back projection with the changed latent code}
\label{fig:image}
\end{figure}\\
Thus, we actually have an autoencoder that is responsible for a new “cube” concept on the second level of processing. In practice, it is a very important mechanism in recognition tasks. Abstract concepts on higher levels allows addressing of ambiguities.

\section{Conclusions}

A mathematical approach for estimation of likelihood function of different models that describes the input data is proposed. Models were assumed to be autoencoders. The following ideas concerning these autoencoders  are proposed: their latent space can be shared and that latent codes correspond to treatment of information. It is also defined that autoencoders or latent models are by itself contexts of that treatment.\\

It was shown that autoencoders are equal to or better than fully-connected neural networks for recognition tasks on the MNIST dataset.

An effect of “one-shot learning” is demonstrated. It is based on separating of treatment and context and on transfer learning. 

Ability of forming new concepts is shown on a sample where a 3D cube concept was received in two-level processing scheme.

\section{Source codes}

Link to all source codes of experiments described in the article: \url{https://gitlab.com/Morzhakov/SharedLatentSpace}

\end{document}